# scientific reports

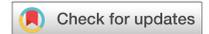

OPEN

# Significantly improving zero-shot X-ray pathology classification via fine-tuning pre-trained image-text encoders

Jongseong Jang[1,3], Daeun Kyung[2,3], Seung Hwan Kim[1], Honglak Lee[1], Kyunghoon Bae[1] & Edward Choi[2✉]

Deep neural networks are increasingly used in medical imaging for tasks such as pathological classification, but they face challenges due to the scarcity of high-quality, expert-labeled training data. Recent efforts have utilized pre-trained contrastive image-text models like CLIP, adapting them for medical use by fine-tuning the model with chest X-ray images and corresponding reports for zero-shot pathology classification, thus eliminating the need for pathology-specific annotations. However, most studies continue to use the same contrastive learning objectives as in the general domain, overlooking the multi-labeled nature of medical image-report pairs. In this paper, we propose a new fine-tuning strategy that includes positive-pair loss relaxation and random sentence sampling. We aim to improve the performance of zero-shot pathology classification without relying on external knowledge. Our method can be applied to any pre-trained contrastive image-text encoder and easily transferred to out-of-domain datasets without further training, as it does not use external data. Our approach consistently improves overall zero-shot pathology classification across four chest X-ray datasets and three pre-trained models, with an average macro AUROC increase of 4.3%. Additionally, our method outperforms the state-of-the-art and marginally surpasses board-certified radiologists in zero-shot classification for the five competition pathologies in the CheXpert dataset.

The success of deep neural networks (DNNs) for visual and textual data led to their wide adoption in diverse domains and tasks, including pathology classification based on medical images. Although there have been some successful applications of neural networks to medical images, such as diabetic retinopathy classification[1,2] and skin cancer detection[3], their widespread adoption has been hindered due to the need for large amounts of high-quality training samples. This poses a great challenge especially in the medical imaging domain, as qualified experts are required to create labeled samples, not to mention the difficulty of collecting images to begin with as they require special machinery. Recent studies tackled this problem with some success by leveraging advantage of the powerful pre-trained models trained on large-scale general domain data. Some demonstrated that fine-tuning a pre-trained image encoder in a supervised manner can yield impressive performance[4–6]. However, this approach still requires a certain amount of expert-annotated images. To mitigate this dependency, researchers are exploring medical image-text representation learning[7–10], inspired by large-scale image-text pre-training models like CLIP[11]. For instance, ConVIRT[7] introduces contrastive learning to the medical domain, GloRIA[8] aims to align image and text embeddings both globally and locally. BioViL[9] incorporates temporal content by accounting for prior images and reports. CheXzero[10], closely related to our approach, demonstrates that effective pathology classification can be achieved without an expert-annotated training set by aligning X-ray images with reports from the MIMIC-CXR dataset, in the same manner as CLIP.

However, the models often fail to properly address the false-negative case, i.e., image-report pairs with the same label from different patients. This issue becomes evident when considering the medical image-report pairs with their multi-labeled nature. For example, let us imagine two chest X-ray studies where one has Lung Opacity, Edema, Pneumonia, Support Devices, and the other has the first two. Ideally, their reports should share the semantics of Lung Opacity and Edema (Fig. 1). However, most of medical contrastive learning methods may inaccurately categorize these partially overlapping pairs as entirely distinct, leading to sub-optimal representation learning. While MedCLIP[12] attempts to address this using CheXpert[13] labels, it is constrained by the labeler's limitations and the predefined disease classes. Our method employs loss relaxation, which clips the loss based on similarity between the target sample, thus avoiding reliance on expert labels.

[1]LG AI Research, 07796 Seoul, Republic of Korea. [2]KAIST, Kim Jaechul Graduate School of AI, 34141 Daejeon, Republic of Korea. [3]These authors contributed equally to this work. ✉email: edwardchoi@kaist.ac.kr

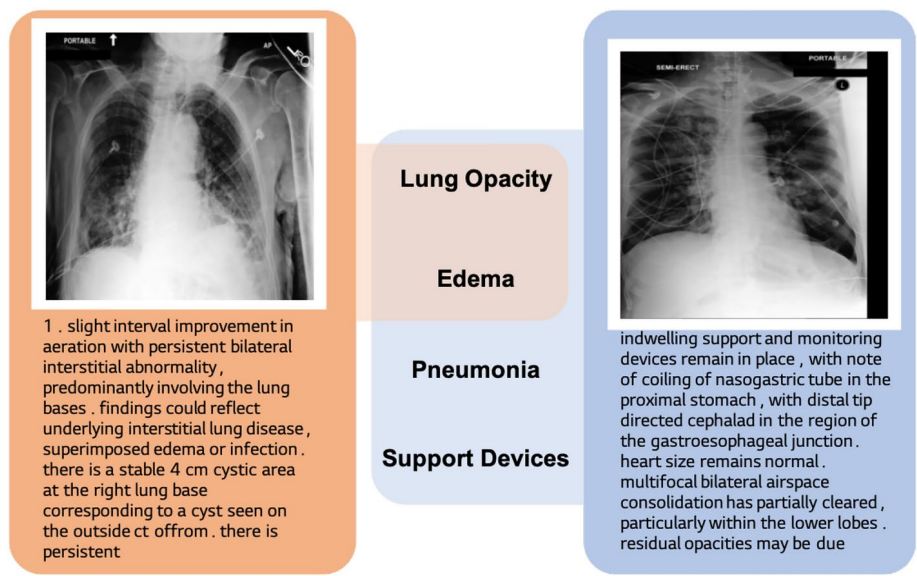

**Figure 1.** Example of false-negatives between medical data. Two image-text pairs share two classes of disease; lung opacity and edema. Existing image-text contrastive learning frameworks deal other pair as perfectly negative one.

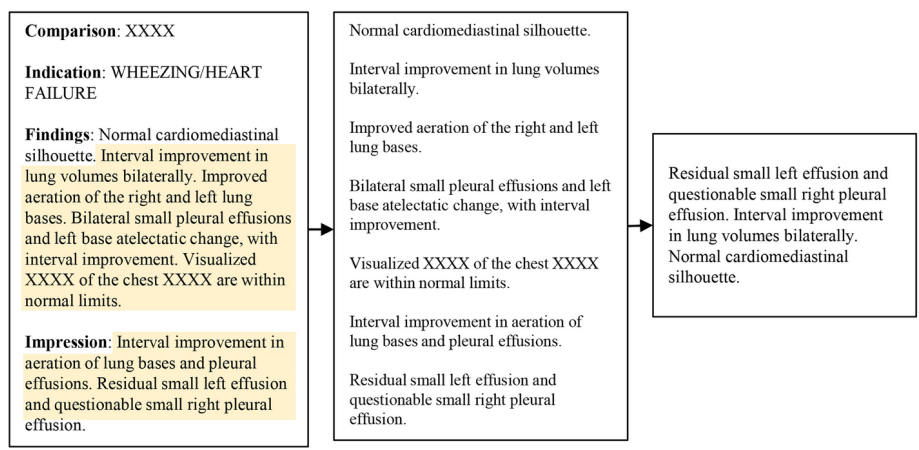

**Figure 2.** Example of random text sampling. Given the original free-form report (left), we extract only the "Findings" and "Impression" section and split them into sentences (middle). $n$ sentences are randomly selected to make new positive text pairs (right).

On the other hand, we propose a new text processing method for medical report to improve the contrastive learning performance. In the medical domain, text augmentation techniques, such as random insertion, deletion, or backtranslation[14] are typically avoided due to the high risk of information loss, as medical reports are often brief and dense with critical information. Previous method[15] involves substituting medical concepts with the corresponding terms of the Unified Medical Language System (UMLS), require both powerful extraction tools and extensive medical knowledge. In contrast, our approach utilizes a straightforward yet effective technique: sentence-level random selection that preserves the content of each sentence as it is, without additional knowledge.

In this work, we propose a new fine-tuning strategy for zero-shot pathology classification by taking into account the multi-label nature of medical image-text pairs. Our method consists of sentence sampling and loss relaxation, which are based on two core observations: (1) Every sentence in a report contains important clinical information; (2) There are many false-negative pairs, but practically no perfectly positive pairs. Our label-free method can be applied to any pre-trained contrastive image-text encoders to enhance downstream zero-shot performance. This fine-tuning strategy consistently improved zero-shot pathology classification when applied to three pre-trained models on four different chest X-ray datasets, even surpassing board-certified radiologists in some instances.

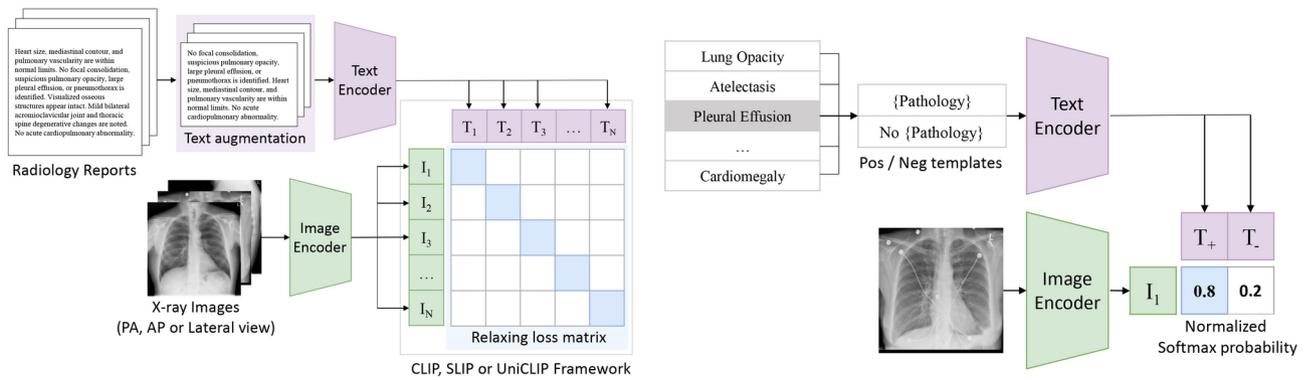

**Figure 3.** (left) Training phase. Our fine-tuning strategies, random text sampling and relaxation loss, on contrastive learning frameworks with a medical multi-modal dataset. Our proposed methods could be easily applied to various image-text contrastive learning frameworks, such as CLIP, SLIP, and UniCLIP. (right) Inference phase. For the zero-shot pathology classification task, the text encoder works as a multi-label classifier by embedding the names or descriptions of the pathologies in the target dataset at test time.

|  | CheXpert | | Open-i | | PadChest | VinDr-CXR |
|---|---|---|---|---|---|---|
|  | Avg. 5 AUC | Avg. total AUC | Avg. 5 AUC | Avg. total AUC | Avg. total AUC | Avg. total AUC |
| CLIP | 0.870 ± 0.006 | 0.771 ± 0.001 | 0.785 ± 0.030 | 0.691 ± 0.010 | 0.685 ± 0.026 | 0.801 ± 0.020 |
| CLIP w/ ours | **0.885 ± 0.007** | **0.789 ± 0.017** | **0.808 ± 0.008** | **0.708 ± 0.013** | **0.725 ± 0.014** | **0.842 ± 0.007** |
| SLIP | 0.831 ± 0.008 | 0.722 ± 0.008 | 0.760 ± 0.008 | 0.666 ± 0.004 | 0.684 ± 0.004 | 0.799 ± 0.018 |
| SLIP w/ ours | **0.884 ± 0.009** | **0.779 ± 0.020** | **0.810 ± 0.016** | **0.719 ± 0.018** | **0.741 ± 0.005** | **0.831 ± 0.009** |
| UniCLIP | 0.851 ± 0.008 | 0.741 ± 0.037 | 0.744 ± 0.032 | 0.643 ± 0.027 | 0.684 ± 0.009 | 0.759 ± 0.019 |
| UniCLIP w/ ours | **0.888 ± 0.006** | **0.802 ± 0.018** | **0.788 ± 0.016** | **0.694 ± 0.006** | **0.717 ± 0.011** | **0.843 ± 0.016** |

**Table 1.** The zero-shot classification performance of three different CLIP-based frameworks across four distinct datasets is provided (mean ± std). The term 'w/ ours' refers to models that have been fine-tuned with our strategy, including random sentence sampling and loss relaxation. **Bold** means the best performance within each framework. For the CheXpert and Open-i datasets, we report the average AUC of 5 CheXpert competition pathologies and a total of 13 pathologies, respectively. For the PadChest and VinDr-CXR datasets, the average of 61 pathologies (where each has more than 100 samples) and 20 pathologies (where each has more than 20 samples), respectively.

The contribution of this work is summarized below:

- We propose a new fine-tuning strategy, random sentence sampling and loss relaxation, for zero-shot pathology classification from unlabeled X-ray images. Our methods can be applied easily across various pre-trained contrastive image-text encoders without the need for external knowledge or architectural modifications.
- We demonstrate that our proposed method consistently improves zero-shot pathology classification performance across four different chest X-ray datasets and three distinct pre-trained models. Notably, for the CheXpert dataset, our method sometimes outperforms board-certified radiologists in detecting five key diseases.
- Our fine-tuning strategy also showed consistent performance improvement for image-text alignment. Ablation studies were conducted simultaneously to verify the effectiveness of the proposed fine-tuning method.

## Methods and materials
### Zero-shot X-ray pathology classification

In this paper, we focus on the zero-shot X-ray pathology classification task following CheXzero[10]. Zero-shot evaluation refers to the process of training or fine-tuning a model on a large-scale dataset and then testing its performance on different datasets without further fine-tuning, known as cross-dataset zero-shot evaluation[11,16]. We specifically address this challenge within the medical domain, using chest X-ray (CXR) images and their corresponding report. To achieve this, we first jointly train an image encoder $E_{img}$ and a text encoder $E_{txt}$ to predict the correct image-text pairs through InfoNCE loss (Fig. 3 (left)). Notably, we do not use manual labels for X-ray images during training. For the inference phase, we use the similarity between the image embedding of the target unseen X-ray image $\mathbf{I}_{trg} \in \mathbb{R}^{H \times W \times 3}$ and the text embedding of the prompts $\mathbf{T}_c$ for target pathology $c$ to conduct the zero-shot classification. We construct positive and negative prompts, $\mathbf{T}_{+,c}$ and $\mathbf{T}_{-,c}$, for each pathology $c$. We use the prompts of "{label}" and "no {label}" following Tiu et al.[10] (e.g., "Atelectasis" and "No atelectasis"). Then we calculate the softmax probabilities for each pathology based on the cosine similarity between the embedding of the target image $E_{img}(\mathbf{I}_{trg}) \in \mathbb{R}^{d_I}$ and the embedding of the positive/negative prompt

| | mean AUC | Mean F1 | Mean MCC |
|---|---|---|---|
| Radiologists (mean)[10] | - | 0.619 | 0.530 |
| CLIP | 0.870 | 0.578 | 0.423 |
| CLIP w/ ours | 0.885 | 0.603 | 0.443 |
| CLIP$_{ensemble}$ | 0.887 | 0.613 | 0.519 |
| CLIP$_{ensemble}$ w/ ours | 0.893 | 0.614 | 0.525 |
| SLIP | 0.831 | 0.578 | 0.423 |
| SLIP w/ ours | 0.884 | 0.608 | 0.424 |
| SLIP$_{ensemble}$ | 0.845 | 0.564 | 0.456 |
| SLIP$_{ensemble}$ w/ ours | 0.893 | **0.625** | 0.537 |
| UniCLIP | 0.851 | 0.561 | 0.458 |
| UniCLIP w/ ours | 0.888 | 0.610 | 0.522 |
| UniCLIP$_{ensemble}$ | 0.881 | 0.614 | 0.524 |
| UniCLIP$_{ensemble}$ w/ ours | **0.900** | 0.623 | **0.544** |

**Table 2.** Comparison with the performance of three board-certified radiologists on the CheXpert test dataset. We also compare the ensemble over the best model checkpoints from 3 runs with a different random seed. We report the average results across the five CheXpert competition pathologies. The mean AUC of radiologists is not available, as radiologist predictions are only binary. The colored cell means that the model outperformed the prediction performance of the expert radiologist. The best performance is in bold, and the second best is in underlined.

| Testset | # of label | Metric | BioViL-T | CheXzero | Our$_{single,best}$ |
|---|---|---|---|---|---|
| CheXpert | 5 | AUC | 0.812 | 0.844 ± 0.008 | **0.884 ± 0.009** |
| | 5 | F1 | 0.544 | 0.559 ± 0.005 | **0.608 ± 0.018** |
| | 5 | MCC | 0.361 | 0.417 ± 0.017 | **0.424 ± 0.035** |
| | 13 | AUC | 0.779 | 0.733 ± 0.008 | **0.779 ± 0.020** |
| Open-i | 5 | AUC | 0.785 | 0.771 ± 0.006 | **0.810 ± 0.016** |
| | 13 | AUC | 0.714 | 0.684 ± 0.006 | **0.719 ± 0.018** |
| PadChest | 61 | AUC | 0.715 | 0.702 ± 0.011 | **0.741 ± 0.005** |
| VinDr-CXR | 20 | AUC | 0.774 | 0.771 ± 0.024 | **0.831 ± 0.009** |

**Table 3.** Zero-shot classification performance of BioViL-T, CheXzero and our single best model Our$_{single,best}$ (SLIP w/ ours) on the four different datasets (mean ± std). The best performance is in bold. The dataset setting is the same as in Table 1. All AUCs are mean values among diseases.

| | mean AUC | Mean F1 | Mean MCC |
|---|---|---|---|
| Our$_{ensemble,best}$ | **0.900 (0.868, 0.929)** | **0.638 (0.566, 0.707)** | **0.552 (0.468, 0.633)** |
| CheXzero$_{ensemble}$ | 0.824 (0.777, 0.866)[a] | 0.573 (0.493, 0.652)[a] | 0.432 (0.339, 0.528)[a] |
| BioViL | 0.812 (0.762, 0.857)[a] | 0.558 (0.480, 0.634)[a] | 0.423 (0.331, 0.511)[a] |
| Xplainer[b] | 0.864 (0.823, 0.901)[a] | 0.607 (0.531, 0.683)[a] | 0.493 (0.403, 0.585)[a] |

**Table 4.** Zero-shot classification performance of CheXzero, BioViL-T, Xplainer, and our best ensemble model, Our$_{ensemble, best}$ (UniCLIP$_{ensemble}$ w/ ours), on the CheXpert test set using bootstrap for five competition pathologies (Atelectasis, Cardiomegaly, Consolidation, Edema, Pleural Effusion). The numbers in parentheses indicate the 95% confidence intervals (CI). [a]p-value < 0.001, Our$_{ensemble,best}$ wins significantly. [b]Use external observation descriptions.

|  | Open-i | | PadChest | |
|---|---|---|---|---|
|  | Sentence-level | Report-level | Sentence-level | Report-level |
| CLIP | 0.275 ± 0.053 | 0.407 ± 0.007 | 0.102 ± 0.099 | 0.135 ± 0.111 |
| CLIP w/ours | 0.658 ± 0.005 | 0.693 ± 0.005 | 0.572 ± 0.021 | 0.579 ± 0.015 |
| SLIP | 0.163 ± 0.021 | 0.287 ± 0.031 | 0.007 ± 0.022 | 0.018 ± 0.014 |
| SLIP w/ours | 0.434 ± 0.034 | 0.489 ± 0.035 | 0.512 ± 0.008 | 0.500 ± 0.009 |
| UniCLIP | 0.209 ± 0.013 | 0.318 ± 0.016 | 0.096 ± 0.010 | 0.118 ± 0.011 |
| UniCLIP w/ours | 0.451 ± 0.038 | 0.543 ± 0.024 | 0.396 ± 0.060 | 0.418 ± 0.068 |
| CheXzero | 0.319 ± 0.015 | 0.387 ± 0.015 | 0.232 ± 0.079 | 0.258 ± 0.081 |
| BioViL-T | 0.628 | 0.315 | 0.241 | 0.158 |

**Table 5.** Image-text similarity for Open-i and PadChest datasets (mean ± std). We report both sentence-level similarity and report-level similarity for each dataset.

|  | Text aug | Relax loss | Atelectasis | Cardiomegaly | Consolidation | Edema | Pleural effusion | Avg. 5 AUC | Avg. total AUC |
|---|---|---|---|---|---|---|---|---|---|
| CLIP |  |  | 0.828 ± 0.016 | 0.845 ± 0.012 | 0.895 ± 0.007 | 0.861 ± 0.016 | 0.878 ± 0.020 | 0.861 ± 0.002 | 0.787 ± 0.044 |
|  |  | ✓ | 0.858 ± 0.009 | 0.815 ± 0.013 | 0.889 ± 0.023 | 0.875 ± 0.022 | **0.890 ± 0.006** | 0.865 ± 0.002 | 0.794 ± 0.045 |
|  | ✓ |  | **0.864 ± 0.004** | **0.867 ± 0.008** | 0.904 ± 0.015 | **0.903 ± 0.003** | 0.877 ± 0.021 | **0.883 ± 0.005** | <u>0.820 ± 0.001</u> |
|  | ✓ | ✓ | 0.859 ± 0.006 | 0.840 ± 0.016 | **0.913 ± 0.010** | 0.881 ± 0.009 | 0.869 ± 0.012 | <u>0.872 ± 0.005</u> | **0.824 ± 0.003** |
| SLIP |  |  | 0.760 ± 0.046 | 0.788 ± 0.008 | 0.877 ± 0.028 | 0.847 ± 0.021 | **0.902 ± 0.010** | 0.835 ± 0.010 | 0.791 ± 0.015 |
|  |  | ✓ | 0.826 ± 0.010 | 0.828 ± 0.010 | 0.899 ± 0.006 | 0.873 ± 0.013 | 0.899 ± 0.010 | <u>0.865 ± 0.005</u> | **0.828 ± 0.003** |
|  | ✓ |  | 0.830 ± 0.022 | 0.837 ± 0.021 | 0.864 ± 0.042 | 0.899 ± 0.008 | 0.876 ± 0.013 | 0.861 ± 0.004 | 0.800 ± 0.015 |
|  | ✓ | ✓ | **0.845 ± 0.014** | **0.855 ± 0.018** | **0.904 ± 0.014** | **0.905 ± 0.016** | 0.868 ± 0.009 | **0.875 ± 0.006** | <u>0.819 ± 0.028</u> |
| UniCLIP |  |  | 0.774 ± 0.060 | 0.833 ± 0.012 | 0.869 ± 0.016 | 0.856 ± 0.033 | 0.882 ± 0.020 | 0.843 ± 0.010 | 0.750 ± 0.016 |
|  |  | ✓ | 0.840 ± 0.009 | 0.818 ± 0.021 | 0.870 ± 0.012 | 0.869 ± 0.014 | **0.897 ± 0.021** | 0.859 ± 0.008 | <u>0.808 ± 0.023</u> |
|  | ✓ |  | 0.860 ± 0.012 | **0.862 ± 0.010** | 0.900 ± 0.019 | 0.879 ± 0.005 | 0.887 ± 0.012 | <u>0.877 ± 0.009</u> | 0.790 ± 0.020 |
|  | ✓ | ✓ | **0.870 ± 0.007** | 0.844 ± 0.007 | **0.911 ± 0.002** | **0.886 ± 0.014** | 0.882 ± 0.017 | **0.878 ± 0.001** | **0.825 ± 0.028** |

**Table 6.** Ablation study of different components of our proposed method on the CheXpert validation set (mean ± std). We report the zero-shot classification performance on three different CLIP-like frameworks. Best performance is in bold and second best is in underline for each framework. For the CheXpert dataset, the AUC for each of the five competition pathologies (Atelectasis, Cardiomegaly, Consolidation, Edema, Pleural Effusion), the average value of these five pathologies, and the total of 13 pathologies of this dataset are reported.

|  | $n=1$ | $n=2$ | $n=3$ | $n=4$ | $n=5$ |
|---|---|---|---|---|---|
| Avg. # of report | 8433.86 | 538.40 | 106.20 | 25.24 | 1.03 |

**Table 7.** Average number of reports which contains same label combinations for CheXpert label. $n$ is number of shared labels.

$E_{txt}(\mathbf{T}_{+,c}), E_{txt}(\mathbf{T}_{+,-}) \in \mathbb{R}^{d_T}$ (Fig. 3 (right)). $d_I$ and $d_T$ means embedding dimension of image encoder and text encoder, respectively.

$$\begin{aligned} s_{+,c} &= E_{img}(\mathbf{I}_{trg}) \cdot E_{txt}(\mathbf{T}_{+,c}), \\ s_{-,c} &= E_{img}(\mathbf{I}_{trg}) \cdot E_{txt}(\mathbf{T}_{-,c}), \\ prob_c &= \text{softmax}(s_{+,c},\ s_{-,c}), \end{aligned} \quad (1)$$

where $c$ indicates target pathology, $s_{+,c}$ and $s_{-,c}$ are the cosine similarity of positive and negative prompt, respectively.

### Random sentence sampling

Unlike the general domain, the medical domain suffers from a scarcity of large-scale, high-quality datasets due to the laborious nature of obtaining expert annotation. Text augmentation is the most widely used solution to address this data shortage. However, the specific nature of medical reports presents unique challenges. Medical reports consist of formal and dense sentences with meaningful content crucial for diagnosis. Traditional augmentation techniques, such as token-level deletion or swapping, risk altering the clinical accuracy by potentially removing essential medical terms (Fig. 2 (left)).

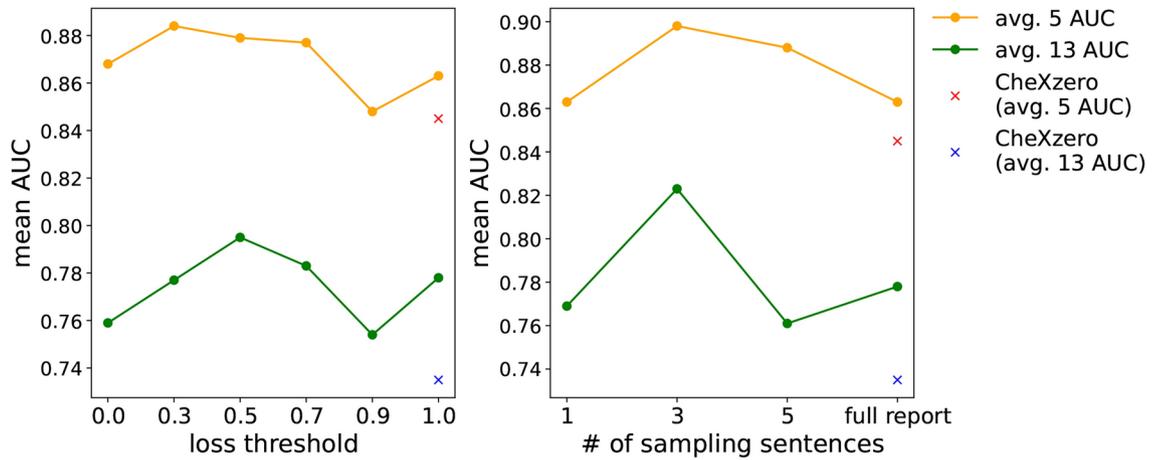

**Figure 4.** Effect of relaxing threshold (left) and number of sampling sentences per report (right). Both evaluate the zero-shot classification performance on the CheXpert validation set.

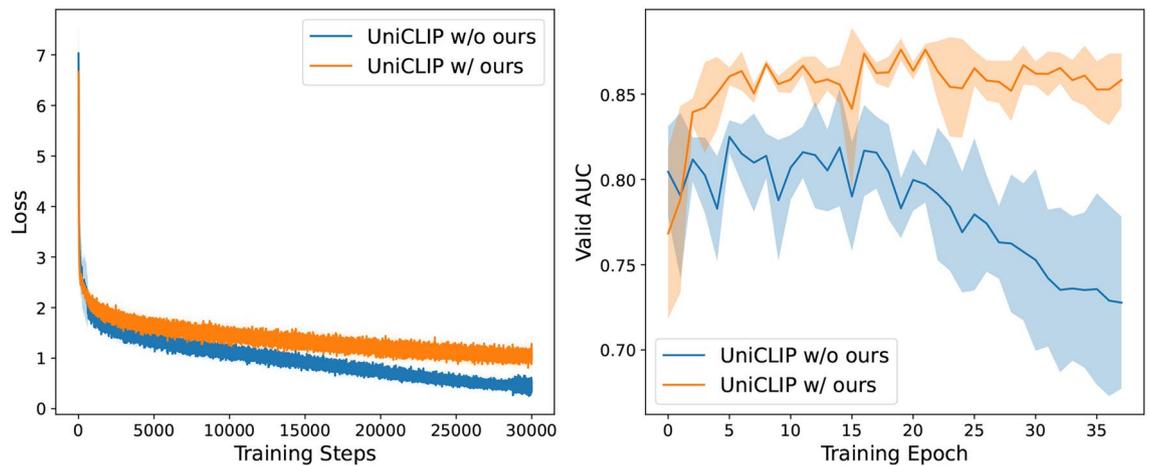

**Figure 5.** Training Loss (left) and AUROC on the CheXpert validation set (right) for two configurations: UniCLIP without our method ('w/o ours') and UniCLIP with our method ('w/ ours').

In addition, given that medical reports resemble lists more than narrative texts, where each sentence corresponds to a specific finding or label, they require specialized processing methods that focus on enhancing embeddings for individual labels, not just the entire report. To address this, we propose a simple, yet effective, unsupervised text processing method to tackle this issue. We randomly sub-sample *n* sentences from the report for every sample and feed them into the contrastive learning framework. Assuming the report contains *m* sentences, the image can be matched to $_mC_n$ positive samples through stochastic sub-sampling. This process enables the model to learn the rich semantics of each sentence, rather than training with just a single positive pair.

### Loss for relaxed image-text agreement
CLIP[11] jointly trains an image encoder and a text encoder to predict the correct pairings of image-text pairs. Given a batch of *N* image-text pairs, CLIP maximizes the similarity between *N* positive pairs while minimizing the similarity between $N^2 - N$ negative pairs by the InfoNCE loss[17], as follows.

$$\mathcal{L} = -\frac{1}{2N}(\sum_{i=1}^{N} log \frac{exp(\mathbf{sim}(u_i, v_i)/\tau)}{\sum_{j=1}^{N} exp(\mathbf{sim}(u_i, v_j)/\tau)} + \sum_{i=1}^{N} log \frac{exp(\mathbf{sim}(v_i, u_i)/\tau)}{\sum_{j=1}^{N} exp(\mathbf{sim}(v_i, u_j)/\tau)}) \qquad (2)$$

where *u, v* are the normalized vectors from the image and text encoders and $(u_i, v_i)$ is a positive pair. **sim** is a function to calculate the similarity between two vectors, and $\tau$ denotes the learnable temperature.

However, the CLIP objective function does not consider false-negative cases, which have similar semantics but are treated as negative pairs. Unfortunately, false-negative pairs (Fig. 1) are more likely to occur due to the fine-grained label scope of the medical domain compared to the general domain. Therefore, enforcing an increase in similarity only for positive pairs, despite the existence of false-negatives, may mislead the model. We argue that reducing the attraction between positive pairs can alleviate this problem. We propose the modified **sim** function that makes the model focus on perfectly negative pairs (images or reports from different patients with different semantics) instead of false-negative pairs by clipping the upper bound of the similarity function.

$$\text{sim}(u_i, v_j) = \begin{cases} \frac{1}{1+\exp(-\alpha(u_i \cdot v_j - t))}, & \text{if } i = j \text{ and } u_i \cdot v_j \geq t \\ u_i \cdot v_j / 2t, & \text{else if } i = j \text{ and } t > u_i \cdot v_j \geq 0 \\ u_i \cdot v_j, & \text{otherwise} \end{cases} \quad (3)$$

, where $u_i \cdot v_j$ denotes cosine similarity between two vectors, $\alpha$ a slope coefficient of the Sigmoid function. A threshold of $t$ ($0 < t < 1$) is used to determine the level of attractiveness between positive pairs. With our modified similarity function, positive pairs with a similarity score $\text{sim}(u_i \cdot v_i) \geq t$ quickly reach a maximum similarity value (i.e., loss of those pairs becomes minimum). Note that we separately defined the second and third terms of (3) for the continuity of the **sim** function when $u_i \cdot v_i = 0$ and $u_i \cdot v_i = t$, ensuring stable backpropagation. Note that CLIP-like methods, such as SLIP and UniCLIP, utilize similarity between samples, despite differences in how they construct the positive set. Therefore, the **sim** function can be easily substituted for those losses.

## Datasets

For model training, we use MIMIC-CXR[18], a large public dataset, as the training set, and evaluate its performance on four different publicly available datasets: CheXpert[13], Open-i[19], PadChest[20], and VinDr-CXR[21]. Each dataset is described in detail below and the statistics for each are presented in Table 8.

**MIMIC-CXR**: MIMIC-CXR[18] is a large-scale public database that includes 377,110 chest radiographs corresponding to 227,835 studies in free-text format. We used this dataset to fine-tune the CLIP frameworks. Since meaningful observations are mainly written in the "Findings" or "Impression" sections, we extract those two sections for each study (https://github.com/MIT-LCP/mimic-cxr). If the section is not extracted by the rule-based method, we use the last paragraph of those samples. All radiographs are used for fine-tuning regardless of view (AP, PA, and lateral).

**CheXpert**: CheXpert[13] is a public dataset from Stanford Hospital, that includes 224,316 chest radiographs of 65,240 patients, annotated with 14 classes. The validation and test sets have 234 and 668 images, respectively. We use CheXpert's validation set for model selection and the test set for evaluation. Since the test and validation sets of the CheXpert dataset are manually annotated by board-certified radiologists, we use the CheXpert validation set for model selection, following the approach used in CheXzero[10]. Specifically, we select the model based on the mean AUROC of the validation set across the five CheXpert competition tasks: Atelectasis, Cardiomegaly, Consolidation, Edema, and Pleural Effusion (https://stanfordmlgroup.github.io/competitions/chexpert). For evaluation, we use not only the above five tasks, but also all labels from the CheXpert test set, except for "no finding." This exception is made since predicting all labels as zero is equivalent to a "no finding" label in a multi-label setting.

**PadChest**: The PadChest[20] is a public dataset provided by the Medical Imaging Databank of the Valencia Region. This dataset consists of more than 160,000 chest radiograph images of 67,000 patients from San Juan Hospital (Spain) from 2009 to 2017. PadChest is annotated with 174 different findings and 19 differential diagnoses. We used only 15,091 examples ($n = 15,091$) as a test set, which board-certified radiologists manually annotated. We used 61 radiographic findings with more than 100 samples in the test set. We use the PadChest test set to evaluate zero-shot classification performance and image-text alignment.

**Open-i**: Open-i collection[22] is a service for abstract and image search by the National Library of Medicine (NLM). The Open-i chest radiograph dataset includes 7470 images and 3,955 reports provided by Indiana University Hospital. We select the "Finding" and "Impression" sections which contain essential information associated with the X-ray images. We first select the image-text pair with frontal view images and filtered samples with a report length of 10 characters or more. We used a pre-processed dataset with 2,550 image-text pairs as our test set to evaluate image-text similarity and zero-shot classification. For reference, the data is labeled using the CheXpert labeler tool provided by the authors (https://github.com/stanfordmlgroup/chexpert-labeler)[13].

| | | # of Images | # of Reports | # of classes in use |
|---|---|---|---|---|
| Fine-tuning | MIMIC-CXR[18] | 377,110 | 227,835 | – |
| | CheXpert$_{valid}$[13] | 234 | – | 5/14 |
| Evaluation | CheXpert$_{test}$[13] | 668 | – | **5, 13**/14 |
| | Open-i[32] | 2,550 | 2,550 | **5, 13**/14 |
| | PadChest[20] | 15,091 | 15,091 | **61**/174 |
| | VinDr-CXR[21] | 3,000 | – | **20**/28 |

**Table 8.** The statistics of datasets.

**VinDr-CXR**: VinDr-CXR[21] includes 18,000 images, 15,000 for the train set and 3,000 for the test set, collected from two major hospitals in Vietnam. A total of 17 board-certified radiologists annotated them with 28 labels consisting of 22 local abnormalities and 6 global suspected diseases. Each scan in the test set was labeled by consensus of 5 radiologists. We evaluate zero-shot classification for 20 labels with more than 20 samples within the test set.

### Implementation details

In this paper, we fine-tune three CLIP-based frameworks, CLIP[11], SLIP[23] and UniCLIP[24] with and without our proposed methods to demonstrate their effectiveness. To evaluate the efficacy of our approach, we conducted comparisons against several recently proposed state-of-the-art, label-free methods, CheXzero[10] and BioViL-T[9]. Additionally, we compared our method to Xplainer[25], which integrates external knowledge about target diagnoses with the BioViL framework. It is important to note that Xplainer requires additional descriptions for each target diagnosis, limiting its application to only the CheXpert dataset among our target test datasets. Our CLIP framework consists of ViT-B/16 and $\text{BERT}_{\text{base}}$ without our proposed method, which might appear similar to the CheXzero configuration. However, CheXzero uses ViT-B/32 as an image encoder and $\text{BERT}_{\text{base}}$ as a text encoder. The baseline "CheXzero" in this paper follows the original configuration for fair evaluation. In all settings, the dimensions for the image and text encoders, $d_I$ and $d_T$, are consistently fixed at 768.

We utilized the $\text{BERT}_{\text{base}}$ model[26] as our text encoder and the ViT-B/16[27] as our image encoder within three CLIP-based frameworks as described above (i.e., CLIP, SLIP, UniCLIP). For CLIP and UniCLIP, we initialize the model using the publicly available pre-trained weight provided by OpenAI (https://github.com/openai/CLIP). In the case of SLIP, we use the model checkpoint that was pre-trained with the YFCC-15M dataset[28] for initialization (https://github.com/facebookresearch/SLIP). Note that all baseline models, including CLIP, SLIP, and UniCLIP, have been fine-tuned following the methods outlined in their respective papers. The term "w/ ours" indicates models that were fine-tuned using their original methods, but with the addition of our strategy, which includes random sentence sampling and loss relaxation. For BioViL-T and Xplainer, we used the officially released checkpoints from Hugging Face (https://huggingface.co/microsoft/BiomedVLP-BioViL-T), as neither framework provided training scripts. For CheXzero, we retrained the model using our training hyperparameters and image augmentation settings, as we found that our configuration achieved a higher CheXpert validation AUROC (0.870 vs. 0.864) than the original configuration.

In our text sampling approach, we randomly sampled three sentences ($n = 3$) from each report. The relaxation loss threshold was set at $t = 0.5$, and the slope coefficient $\alpha$ was fixed at 10. For image augmentation, we apply random affine transformations with the following hyperparameters, rotations: (− 20, 20), resize crop ratio: (0.9, 1.1), horizontal flip: true/false, color jitter (brightness and contrast): (0.5, 2). These augmentations are based on the UniCLIP setting[24], but we reduced their intensity to avoid cropping important parts of the CXR. While CLIP and SLIP primarily use random square cropping from resized images during the pre-training phase, we adopted the random affine image augmentation method for fine-tuning since we found that these augmentations improved the CheXpert validation AUROC. We set the image resolution at $224 \times 224$. We implement the entire model in PyTorch[29] and train the network over 50 epochs with a batch size of 1024 across 8 GPUs on the A100 server (NVIDIA Crop.). We use the Adam optimizer[30] with an initial learning rate $1 \times 10^{-4}$ and decay the learning rate using a cosine scheduler with a warm-up for 100 iterations[31]. We conducted a hyperparameter sweep over batch size and learning rate using the CheXpert validation dataset. We first selected the hyperparameters for CLIP fine-tuning without our method, then added our method while keeping all other parameters unchanged. This demonstrates the advantage of our method, as it can improve model performance simply by adding it without altering any other parameters. To ensure robustness, we repeat all experiments three times with different random seeds and report the mean and standard deviation for the single model results. The ensemble results are derived from the average predictions of these three models. Since standard deviations cannot be reported for the ensemble model, we include confidence intervals and p-values to demonstrate statistical significance. These statistics are calculated using a non-parametric bootstrap method, as detailed in the revised version of CheXzero. Specifically, we generated 1,000 bootstrap samples of size *n* (equal to the size of the original dataset) with replacement.

### Evaluation tasks

*Zero-shot classification*
We perform a zero-shot multi-label image classification on four datasets; CheXpert, PadChest, Open-i, and VinDr-CXR. For CheXpert and Open-i, we use all 13 pathologies. For PadChest, we use 61 labels out of a total of 193, each of which has more than 100 patients. Finally, VinDR-CXR is evaluated with 20 of 28 labels that include more than 20 patients. The macro AUROC (Area under the ROC curve) for multi-label is used as an evaluation metric. We calculated the binary AUROC for each label separately and then computed the macro average.

*Image-text alignment*
In a medical report, each sentence generally contains different label information. Therefore, measuring image-text alignment not only at the report level but also at the sentence level can indicate how the learned embedding contains semantics for each label. We calculate the cosine similarity between image embedding and report embedding for each X-ray-Report pair for report-level evaluation. We measure sentence-level alignment by calculating the average similarity between the image and the sentence embedding for all sentences in a given report.

## Results
### Zero-shot classification

We empirically validate that our fine-tuning strategies consistently improve the zero-shot classification performance on 4 independent datasets collected from four different countries: CheXpert[13], Open-i[19], PadChest[20], and VinDR-CXR[21]. CheXpert and Open-i are evaluated on the same CheXpert 13 labels. PadChest and VinDR-CXR are evaluated on 61 labels and 20 labels, respectively. A list of all labels used for evaluation is in the Table 9. Please refer to the "Datasets" section in the Method for detailed information about those evaluation datasets.

To analyze the effect of our proposed method, we experimented with three different CLIP-like frameworks: CLIP[11], SLIP[23] and UniCLIP[24]. In Table 1, we demonstrate the effectiveness and generalizability of our method by showing consistent results in different labels and data distributions. We show significant improvements compared to baseline performance (i.e., using our fine-tuning strategy) in the three CLIP-like frameworks and four datasets. We observe an average 5.77% increase in AUROC is zero-shot classification. For details, the CLIP baseline improved by 3.4% increase, SLIP by 6.86%, and UniCLIP by 7.05%. These results mean that our method successfully embeds the multi-pathology feature, reflecting the nature of the medical data well. In addition, we also show that using an ensemble of selected models brings significant improvements in all cases (Table 2). We also show that the model can perform pathology classification performance comparable to that of board certified radiologists using our method in Table 2. Furthermore, in the case of SLIP and UniCLIP, they outperform the expert-level by adopting our strategy with ensembles.

We compare our best single results, achieved with SLIP, to the current state-of-the-art models in Table 3. Our model consistently outperforms CheXzero across all datasets and surpasses BioViL-T in most metrics. The only exception is the total AUROC for the CheXpert dataset, where our model's performance aligns with BioViL-T. These findings confirm that our approach has the potential to deliver superior performance relative to existing state-of-the-art models. In Table 4, we present the mean performance of our best ensemble model (UniCLIP$_{ensemble}$ w/ ours) alongside the current state-of-the-art models, detailing their performance differences with a 95% confidence interval (CI). Our model also significantly exceeds the performance of existing models on all reported metrics for the zero-shot classification of the CheXpert dataset.

### Image-text alignment

We analyze how our method enhances the semantic representation quality of the model embeddings by improving image-text alignment at both the sentence and the report levels. To elaborate, "sentence-level" alignment targets individual sentences within a report, whereas "report-level" alignment considers the report's entire content. Given that each sentence in a medical report often presents unique clinical findings, we hypothesize that sentence-level alignment assessment provides insights into the model's ability to handle detailed multi-label information.

The effectiveness of our approach is evidenced by significant improvements in model alignment at both the sentence and report levels, as shown in Table 5. Our method also narrows the disparity between these two levels of measurement. BioViL-T[9], which employs sentence shuffling during text encoder pre-training, shows higher sentence-level similarity than the CLIP-base framework without our strategy. However, a significant gap in report-level similarity remains with these baselines. Our random text sampling prevents the model from overfitting to specific report sections. This approach not only enables the model to focus on crucial pathological information in individual sentences, but also enhances its ability to represent these insights in semantically rich report-level embeddings. One potential concern with our methodology is the possibility of losing critical information during sub-sampling of sentences. Our training strategy involves randomly sub-sampling three sentences per epoch over 50 epochs. Considering that each MIMIC-CXR report contains seven sentences on average, our method is expected to cover all sentences in most reports at least once. Evidence for this is the higher report-level and sentence-level similarity, even surpassing those trained with entire reports. Furthermore, the succinct and structured nature of medical reports makes our method particularly effective in bridging the domain gap across different datasets. Despite the PadChest reports being in Spanish and having a more significant data gap with our training dataset, MIMIC-CXR, our approach still shows notable performance improvements (Table 5).

| | Names |
|---|---|
| CheXpert[13] | Enlarged cardiomediastinum, cardiomegaly, lung opacity, lung lesion, edema, consolidation, pneumonia, atelectasis, pneumothorax, pleural effusion, pleural other, fracture, support devices |
| Open-i[19] | Enlarged cardiomediastinum, cardiomegaly, lung opacity, lung lesion, edema, consolidation, pneumonia, atelectasis, pneumothorax, pleural effusion, pleural other, fracture, support devices |
| PadChest[20] | Air trapping, alveolar pattern, aortic atheromatosis, aortic button enlargement, aortic elongation, apical pleural thickening, atelectasis, bronchiectasis, bronchovascular markings, bullas, calcified densities, calcified granuloma, callus rib fracture, cardiomegaly, consolidation, copd signs, costophrenic angle blunting, descendent aortic elongation, diaphragmatic eventration, dual chamber device, emphysema, fibrotic band, flattened diaphragm, goiter, granuloma, gynecomastia, heart insufficiency, hemidiaphragm elevation, hiatal hernia, hilar congestion, hilar enlargement, hyperinflated lung, increased density, infiltrates, interstitial pattern, kyphosis, laminar atelectasis, mammary prosthesis, metal, nipple shadow, nodule, osteopenia, osteosynthesis material, pacemaker, pleural effusion, pleural thickening, pneumonia, pseudonodule, pulmonary mass, sclerotic bone lesion, scoliosis, single chamber device, sternotomy, supra aortic elongation, suture material, tracheal shift, tuberculosis sequelae, vascular hilar enlargement, vertebral anterior compression, vertebral degenerative changes, volume loss |
| VinDr-CXR[21] | Aortic enlargement, atelectasis, calcification, cardiomegaly, consolidation, ILD, infiltration, lung opacity, mediastinal shift, nodule/mass, pleural effusion, pleural thickening, pneumothorax, pulmonary fibrosis, rib fracture, other lesion, lung tumor, pneumonia, tuberculosis, other disease |

**Table 9.** All labels used in this paper.

## Discussion

Our method is based on two factors: random sentence sampling within the report and loss relaxation to regularize image-text agreement. We analyze the effectiveness of the factors of our method by comparing the zero-shot pathology classification performance on the CheXpert validation set. In Table 6, we verify the benefits conferred by each component. While loss relaxation deals with potential false-negative cases, random sentence sampling directly enriches the diversity of the training data, encouraging the model to learn a more robust embedding.

Our experimental results, as presented in Table 6, indicate that performance is typically optimal when both strategies are employed. Hence, we suggest using both fine-tuning approaches as a standard solution. The degree of enhancement can vary depending on the pathology. For example, the performance of pleural effusion improved primarily with loss relaxation, while other pathologies benefited more from text sampling. Such disparities could arise from variations in how each pathology is described in reports and differences in the distribution of labels between data.

We determined the optimal hyperparameters *n* and *t* for our strategy by evaluating zero-shot classification performance with various settings, as shown in Fig. 4. Our hyperparameters are deciding number of sentences (*n*) to sample in a report and the threshold (*t*) for transforming cosine similarity during fine-tuning. Based on our observations, we set the sentence sampling hyperparameter to $n = 3$ and the loss relaxation hyperparameter to $t = 0.5$.

Sentence sampling may generate false-negative pairs if randomly extracted sentences unintentionally have the same labels. Table 7 shows that, on average, 8, 433 reports share one label, 538 two labels, etc. From Table 7 and Fig. 4, we can conclude that sampling three sentences per report makes a great compromise between the number of false-negatives and the effectiveness of contrastive learning. Note that there still exists partial overlap between sampled sentences, but this is addressed by our loss relaxation.

In terms of computational load, our method, which includes text sampling and loss relaxation, can be directly applied to existing CLIP-based models without any changes to the model architecture. Therefore, the potential impact on computational load of our method is limited to random text selection and the loss computation process. These operations do not significantly impact the overall computational load, as they are lightweight and efficiently handled by modern deep learning frameworks. Consequently, the time efficiency of our method is equivalent to that of the base models when both are trained over the same number of epochs. To analyze the impact of our method on training efficiency, we provide training loss graphs and validation AUROC graphs for UniCLIP and UniCLIP with our method. As shown in Fig. 5, the difference in the speed of loss reduction is not substantial, indicating that our method does not significantly affect the overall training time needed for saturation. Additionally, UniCLIP w/ ours consistently achieves higher validation AUROC values compared to UniCLIP w/o ours. It reaches higher performance levels more rapidly and maintains stability throughout the training epochs. These results demonstrate that our methodology effectively improves the model performance without introducing significant computational overhead.

In this paper, we propose a simple yet effective fine-tuning strategy for medical image-text multi-modal learning. Our method consists of two ways, sentence sampling and loss relaxation, to reflect the multi-label nature of the medical image-text data. Our fine-tuning strategy improves zero-shot pathology classification performance in all three CLIP-based frameworks. Furthermore, our proposed method significantly outperformed board-certified radiologists without additional label information for two frameworks. Additionally, we validate the effect of our method through image-text alignment tasks on two radiography datasets. Although our model shows comparable, or sometimes superior performance compared to the expert level in pathology classification performance, it is difficult to determine whether the model's judgments are based on the location of the pathology. For future work, we plan to explore the medical multi-modal representation learning method that can consider these localized features.

## Data availability

The datasets analyzed in this study are available on MIMIC-CXR (https://physionet.org/content/mimic-cxr/2.0.0/, https://www.physionet.org/content/mimic-cxr-jpg/2.0.0/), CheXpert (https://stanfordmlgroup.github.io/competitions/chexpert/), Open-i (https://openi.nlm.nih.gov/), PadChest (https://bimcv.cipf.es/bimcv-projects/padchest/), and VinDr-CXR (https://physionet.org/content/vindr-cxr/1.0.0/).

## Acknowledgements
This work was conducted during the internship at LG AI Research. This work was supported by the Institute of Information & Communications Technology Planning & Evaluation (IITP) grant (No.RS-2019-II190075), and the Korea Health Industry Development Institute (KHIDI) grants (No.HI21C1138, No.HR21C0198) funded by the Korea government (MSIT, MOHW).

## Author contributions
J.J. and D.K.: Designed the methodology and analysis, processed the data, performed the experiments and analysis, and wrote the paper. S.K., H.L., K.B.: Provided resources and reviewed the manuscript. E.C.: Provided supervision, analysis, and reviewed and edited the manuscript.

## Additional information
**Correspondence** and requests for materials should be addressed to E.C.

**Reprints and permissions information** is available at www.nature.com/reprints.

**Publisher's note**  Springer Nature remains neutral with regard to jurisdictional claims in published maps and institutional affiliations.